\documentclass{ecai} 

\usepackage{latexsym}
\usepackage{amssymb}
\usepackage{amsmath}
\usepackage{amsthm}
\usepackage{booktabs}
\usepackage{enumitem}
\usepackage{graphicx}
\usepackage{color}
\usepackage{xcolor}
\usepackage{colortbl}
\usepackage{url}
\usepackage{hyperref}

\usepackage{soul} 

\newtheorem{definition}{Definition}

\newcommand{\BibTeX}{B\kern-.05em{\sc i\kern-.025em b}\kern-.08em\TeX}
\definecolor{myred}{rgb}{1,0.78,0.80}
\definecolor{mygreen}{rgb}{0.77,0.93,0.80}
\definecolor{myblue}{rgb}{0.79,0.85,0.97}

\begin{document}


\begin{frontmatter}


\paperid{123} 


\title{Can formal argumentative reasoning enhance LLMs performances?}


\author[A]{\fnms{Federico}~\snm{Castagna}}
\author[A]{\fnms{Isabel}~\snm{Sassoon}}
\author[B]{\fnms{Simon}~\snm{Parsons}}

\address[A]{Brunel University London}
\address[B]{University of Lincoln}


\begin{abstract}
Recent years witnessed significant performance advancements in deep-learning-driven natural language models, with a strong focus on the development and release of Large Language Models (LLMs). These improvements resulted in better quality AI-generated output but rely on resource-expensive training and upgrading of models. Although different studies have proposed a range of techniques to enhance LLMs without retraining, none have considered computational argumentation as an option. This is a missed opportunity since computational argumentation is an intuitive mechanism that formally captures agents' interactions and the information conflict that may arise during such interplays, and so it seems well-suited for boosting the reasoning and conversational abilities of LLMs in a seamless manner. In this paper, we present a pipeline (\emph{MQArgEng}) and preliminary study to evaluate the effect of introducing computational argumentation semantics on the performance of LLMs. Our experiment's goal was to provide a proof-of-concept and a feasibility analysis in order to foster (or deter) future research towards a fully-fledged argumentation engine plugin for LLMs. Exploratory results using the MT-Bench indicate that MQArgEng provides a moderate performance gain in most of the examined topical categories and, as such, show promise and warrant further research. 

\end{abstract}

\end{frontmatter}


\section{Introduction}
Since the introduction of the Transformer technology \cite{vaswani2017attention} in 2017, the field of artificial intelligence has significantly advanced by moving towards a paradigm of `pre-training' and `fine-tuning' development \cite{zhao2023survey} ultimately leading to the establishment of the so-called Large Language Models (LLMs). Indeed, as their name suggests, LLMs are precisely scaled-up versions (from the architecture size or data perspective) of pre-trained models. This increase in dimension entails interesting and unforeseen consequences\footnote{Notice, however, that emergent abilities constitute a controversial topic and some researchers even argue against their existence \cite{schaeffer2023emergent}.} impacting the models' capabilities, such as improved arithmetic, multi-task understanding, and enhanced multi-lingual operations \cite{wei2022emergent}. There is also research attesting how Theory of Mind (ToM), i.e. the (human) aptitude to impute mental state to others, may have spontaneously occurred in LLMs as a byproduct of their training \cite{kosinski2023theory}. 
These models thus seem to be endowed with a rich variety of skills that position them far above simple statistical tools which are proficient in language generation. Nonetheless, different scholars argue that LLMs still lack reasoning skills, logical thinking and writing competencies \cite{mahowald2023dissociating,bang2023multitask,frieder2023mathematical,thorp2023chatgptfun}.
Indeed, according to some recent studies \cite{hammond2023large}, LLMs are essentially unable to capture the role of language beyond statistics (and further scaling them up would not provide any adequate solution to this). As an example, we could consider the understanding these models have about causality: whilst they retain correlations from their training data, these do not always reflect reality. Additionally, LLMs can propose likely assertions but not definitive conclusions, which may also change in different instances \cite{hammond2023large}.

Computational argumentation has become increasingly central as a core study within Artificial Intelligence \cite{bench2007argumentation} given its promising paradigm of modelling reasoning in the presence of conflict and uncertainty. Indeed, the main strength of this approach lies within the natural use of arguments as a means to formalise non-monotonic reasoning, showing how humans handle inconsistent information dialectically. In a nutshell, the idea is
that correct reasoning concerns handling only statements whose stance and embedded data can be defended against any challenges moved by counterarguments.

Proposals of integration between computational argumentation and LLMs have already been outlined in \cite{castagnaexplanation,castagna2024computational}, confirming the suitability of such a combination. Building on these initial ideas, in this paper, we developed a simple pipeline (MQArgEng) to enable testing of the effectiveness of computational argumentation semantics in enhancing LLMs performances. Our experiment goal was to provide a proof-of-concept and a feasibility analysis in order to foster (or deter) future research towards a fully-fledged argumentation engine plugin for LLMs. Preliminary results suggest in favour of the former, underscoring also how further studies can potentially yield greater benefits. 

The research original contributions presented herein are twofold and focus on the questions: \emph{(1) Is it feasible to integrate computational argumentation within the Large Language Models workflow?} \emph{(2) Does this yield enhancements in their performance?} \emph{MQArgEng}, the novel LLM and argumentation pipeline introduced here, aims to address both issues.
The paper is structured as follows. In Section 2, we describe the preliminary notions that ground the subsequent work. Section 3 introduces the approaches and tools we leveraged to structure the pipeline presented in Section 4, whose output is assessed via the MT-Bench and recorded in Section 5. These findings are then discussed in Section 6, whereas Sections 7 and 8 review related research and outline potential future directions before concluding in Section 9.     

\section{Background}
Before delving into the envisaged pipeline, we will briefly delineate the formalism characterising the engine responsible for the LLMs `reasoning augmentation', i.e., computational argumentation. Afterwards, we will detail the open-source LLM we harnessed in our pipeline and the benchmark chosen for the final evaluation.
\subsection{Computational Argumentation}
Human interactions are mostly dialectical in nature and revolve around exchanges of arguments and the dialogues that unfold from such interplays. Arguments convey information and, throughout a dispute, they may conflict with other statements. This means that challenging or supporting one's viewpoint often involves argumentative reasoning. This paramount role is also underpinned by scholars who claim that the function of reasoning is indeed argumentative: \emph{``Reasoning has evolved and persisted mainly because it makes human communication more effective and advantageous''} \cite{mercier2011humans}. Drawing from dialectical resolutions of inconsistent information as they occur in everyday interactions, computational argumentation provides an intuitive approach to formally capture arguments and their semantics in order to approximate human reasoning. Crucial to this is the notion of argumentation frameworks, where arguments are treated as abstract entities rendered as nodes in a graph, and every directed edge connects the conflicting arguments of the network:
\begin{definition}[Abstract AFs~\cite{dung1995acceptability}]
An argumentation framework \emph{(AF)} is a pair: $\mbox{AF} = \langle \mbox{AR}, \mathcal{C} \rangle$ where AR is a set of arguments, and $\mathcal{C}$ is the `attack' binary relation on AR, i.e. $\mathcal{C}$ $\subseteq$ AR $\times$ AR.
\end{definition}
The idea conveyed by this formalism is that correct reasoning is rendered via the acceptability of a statement: an argument is \emph{justified} (acceptable) only if it is defended against any counterarguments.
\begin{definition}[Semantics for Abstract AFs~\cite{dung1995acceptability}]
\label{Dung's semantics}
Let $\mbox{AF}=\langle \mbox{AR}, \mathcal{C} \rangle$, and let $\mathcal{S} \subseteq$ AR be a set of arguments. Let also $($X,Y$) \in \mathcal{C}$ denote the conflict existing between an argument $X$ and its target $Y$:
\begin{itemize}
\item $\mathcal{S}$ is \emph{conflict-free} iff $\forall X, Y \in \mathcal{S}$: $(X, Y) \notin \mathcal{C}$;
\item $X \in$ AR is acceptable w.r.t. $\mathcal{S}$ iff $\forall Y \in$ AR such that $(Y, X) \in \mathcal{C}$: $\exists Z \in \mathcal{S}$ such that $(Z, Y) \in \mathcal{C}$;
\item A conflict-free extension $\mathcal{S}$ is an \emph{admissible} extension iff $X \in \mathcal{S}$ implies $X$ is acceptable w.r.t. $\mathcal{S}$;
\item An admissible extension $\mathcal{S}$ is a \emph{complete} extension iff $\forall X \in$ AR: $X$ is acceptable w.r.t. $\mathcal{S}$ implies $X \in \mathcal{S}$. The minimal complete extension $($with respect to set inclusion$)$ is called the \emph{grounded extension}, whereas a maximal complete extension $($with respect to set inclusion$)$ is called a \emph{preferred extension};
\item A \emph{stable extension} $\mathcal{S}$ is such that iff $\forall Y \in$ AR, if $Y \notin \mathcal{S}$, then $\exists X \in \mathcal{S}$ such that $(X, Y) \in \mathcal{C}$. 
\end{itemize}
\end{definition}
\begin{figure}[ht!]
    \centering
    \includegraphics[width=0.8\linewidth]{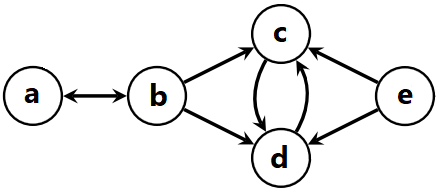}
    \caption{\footnotesize{An abstract argumentation framework.}}
    \label{fig:AFexample}
    \vspace{1cm}
\end{figure}
\noindent Figure \ref{fig:AFexample} provides a graphical example of an AF, its arguments and the conflicting relations existing between them. Following the semantics described in Definition \ref{Dung's semantics}, we can identify the complete (i.e., \{e\}, \{a, e\}, \{b, e\}), grounded (i.e., \{e\}), preferred and stable (i.e., \{a, e\}, \{b, e\}) extensions of the AF.
\subsection{Mistral 7B}
At the time of the design of this experiment, Mistral 7B \cite{jiang2023mistral} was among the most popular and powerful open-source LLMs with seven billion parameters. Engineered by Mistral AI\footnote{\url{https://mistral.ai/}.}, the model hinges upon a Transformer based architecture \cite{vaswani2017attention} characterised by grouped-query attention \cite{ainslie2023gqa} paired with sliding-window attention \cite{beltagy2020longformer} and other memory-saving features (i.e., rolling buffer cache, pre-filling and chunking \cite{jiang2023mistral}). Such an LLM has proved to outperform Llama 2 (7B and 13B \cite{touvron2023llama2}) on a variety of benchmarks and Llama 34B \cite{touvron2023llama} in maths and coding tasks. Given its 8k token context length and its viable inference computational requirement (which makes it accessible also from a consumer-grade laptop), this model seemed a good candidate for our study. However, we needed to run the model in a conversational setting, thus we resorted to the fine-tuned version hosted on the Hugging Face repository\footnote{\url{https://huggingface.co/mistralai/Mistral-7B-Instruct-v0.2}}. This version, i.e., \emph{Mistral-7B-Instruct-v0.2}, improves over the base model by increasing the context length up to 32k tokens (without employing sliding-window attention). 
That being said, discussing an LLM performance is meaningful only when compared with its competitors on specifically designed benchmarks, such as the MT-Bench.

\subsection{MT-Bench}
The crucial role that LLMs are starting to play in our daily lives has rendered the evaluation aspects as pivotal as the training of models, tuning parameters, or development of new heuristics to increase performance. A variety of benchmarks have thus been created to examine LLMs from different perspectives \cite{chang2023survey}. MT-Bench is a multi-turn benchmark that presents 80 challenging queries, divided into two sub-questions each, covering 8 different categories: writing, roleplaying, reasoning, math, coding, extraction, stem and humanities \cite{zheng2023judging}. The idea behind the benchmark is to appraise the conversational capabilities of the tested LLM on a broad range of topics using another LLM as a judge. The evaluator can then focus on one of the following assessment methods (or combine them): \emph{pairwise comparison}, \emph{single answer grading} and \emph{reference-guided grading}. The first approach determines the best among two concurrent models according to their replies to the specific MT-Bench questions. The second assessment involves testing a single model's responses to the aforementioned query by scoring its output in a 1-10 range. On the other hand, reference-guided grading may be employed by both of these two previous approaches when a sample answer would help the evaluation. We decided to appraise our pipeline against the MT-Bench due to the wide array of subjects it accounts for: even in the case that the addition of the argumentation engine plugin will not outscore the base model across all the benchmarks, we would still be able to single out the categories in which it performs better than its contender. Another equally important consideration is that MT-Bench ensures a cheaper and quicker valuation compared to human assessors.
\begin{figure}[h]
    \centering
    \includegraphics[width=1\linewidth]{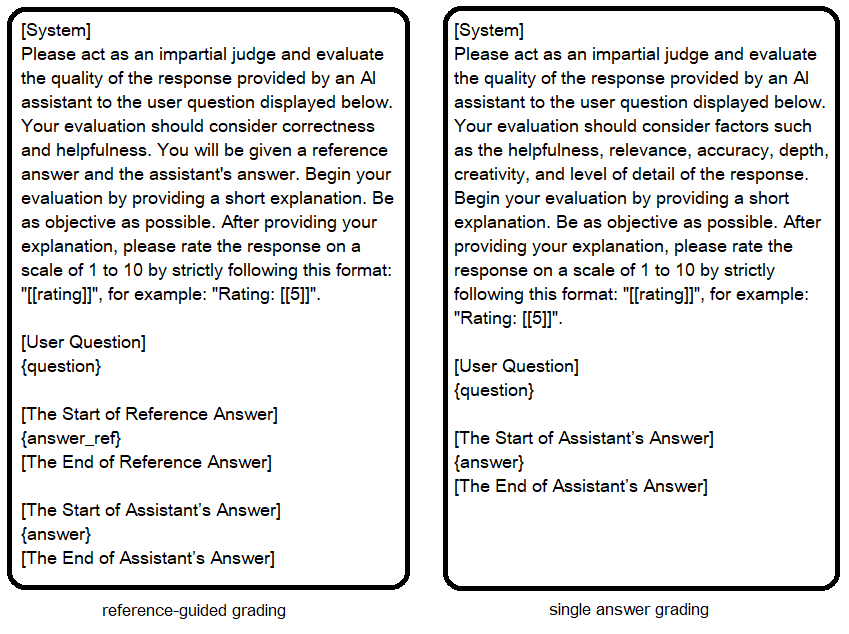}
    \caption{\footnotesize{Examples of MT-Bench prompting templates for single question reference guided (left) and standard (right).}}
    \label{fig:exmaples of MT-Bench prompting1}
    \vspace{1cm}
\end{figure}
\vspace{1cm}
\begin{figure}
    \centering
    \includegraphics[width=1\linewidth]{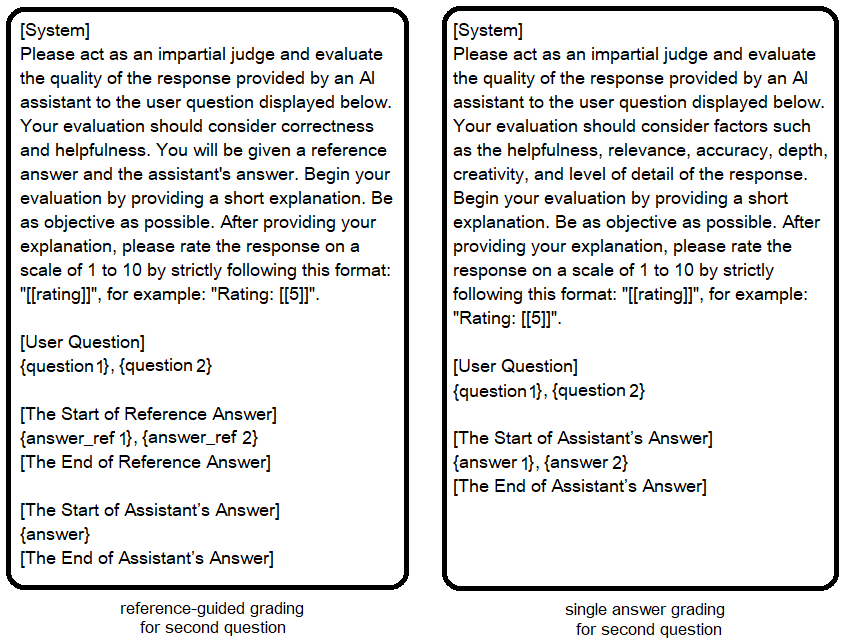}
    \caption{\footnotesize{Examples of MT-Bench prompting templates for multi questions reference guided (left) and standard (right).}}
    \label{fig:exmaples of MT-Bench prompting2}
    \vspace{1cm}
\end{figure}

\section{Methodology}
The pipeline we devised consists of multiple components, the core of which revolves around a plugin. The plugin we engineer can be thought of as an argumentation engine, i.e., a piece of software that leverages the power of computational argumentation to probe the responses of the underlying LLM in order to select the ones that contain acceptable information. The APSARTIX Solver is the tool responsible for identifying arguments and semantics. Additionally, to make the overall process more accessible and less resource-intensive, we quantizated the underlying model. Finally, we designed the final output of the system to be generated from a prompt including a Zero-shot-Chain-of-Thought for a more precise outcome. In the following sections, we briefly outline the mentioned methodologies. 

\subsection{Quantization}
In its essence, quantization is a computational and memory costs-saving approach. Initially described in \cite{jacob2018quantization}, it is based on the idea of representing the model parameters (i.e., weights and activations) in low-precision data types, such as $\mathtt{float16}$ or $\mathtt{int8}$ rather than $\mathtt{float32}$. This procedure usually ensures a substantial reduction in memory storage, energy consumption and yields faster inference \cite{hfquantization}, but it is a trade-off with a (small) loss in accuracy as well. In our experiment, we resort to a quantizated version of Mistral-7B-Instruct-v0.2, henceforth denoted as \emph{MQInstruct}. In particular, we loaded the pre-trained model weights in 4-bit precision to decrease memory footprint by (approximately) 4x times while also enabling nested quantization to preserve further 0.4 bits/parameter. Additionally, to speed up computation, we changed the data type from the default $\mathtt{float32}$ value to $\mathtt{bfloat16}$.

\subsection{ASPARTIX Solver}
\begin{figure}[h!]
    \centering
    \includegraphics[width=0.9\linewidth]{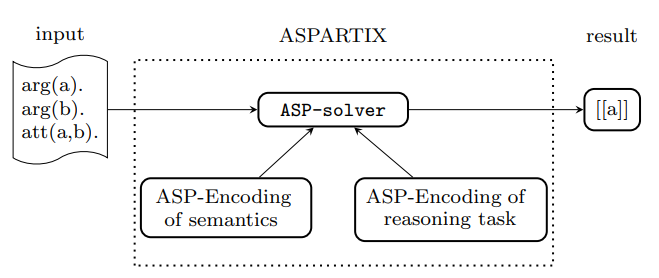}
    \caption{\footnotesize{Overview of the ASPARTIX workflow \cite{dvovrak2020aspartix}.}}
    \label{fig:aspartix-example}
    \vspace{0.5cm}
\end{figure}

The primary goal underpinning computational argumentation theories is to enable the resolution of conflicting knowledge by identifying the most appropriate (i.e. justified) pieces of information. Reinforcing this is the fact that decision-making processes can be encoded as problems whose solutions are rendered by the calculation and evaluation of AFs: an argumentation solver is essentially a reasoning tool driven by the same logic. Such an argumentative decision-making apparatus can be a useful addition to any software application concerning defeasible reasoning, as advocated by the comprehensive study of Bryant and Krause \cite{bryant2008review}. In our pipeline, we leverage the ASPARTIX tool (first introduced in \cite{egly2008aspartix}) as a driving component of the argumentation engine plugin. Its inner workings, summarized in Figure \ref{fig:aspartix-example}, consist of an Answer-Set-Programming (ASP) solver whose input comprises a batch of ASP-Encodings prescribing the specific computation to be applied on the input AF to achieve the required output.

\subsection{CoT}
In order to enhance LLMs' logical thinking without resorting to unceasing updates or retraining of the models, a number of strategies were introduced by the AI community. Chain of Thought (CoT), probably the most influential of such propositions, consists of a prompting technique that details a series of intermediate reasoning steps to achieve better performance in arithmetic, symbolic and commonsense inferences \cite{wei2022chain}. Zero-shot-CoT is a simplified version of CoT that is task-agnostic and does not require few-shot examples \cite{kojima2022large}. To work properly, this strategy necessitates a double prompt comprising: extracting the rationale and extracting the actual answer based on such an uncovered rationale. The first prompt of this approach, rendered by the standard template \emph{``Let's think step by step''}, will also be a component of the final input conveyed into our pipeline. 

\section{MQArgEng} 
\begin{figure}[h!]
    \centering
\includegraphics[width=0.9\linewidth]{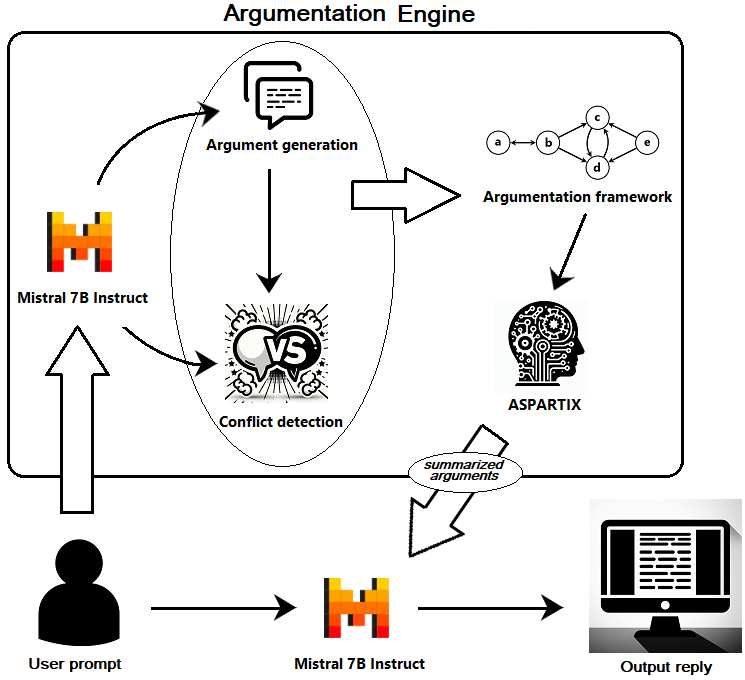}
    \caption{\footnotesize{MQArgEng: Naive pipeline employing the argumentation engine.}}
    \label{fig:naive pipeline}
    \vspace{1cm}
\end{figure}

\emph{MQArgEng} is the pipeline that we devised by leveraging MQInstruct as the underlying model and an argumentation engine as a plugin software in charge of guiding the overall workflow. The pipeline is naive in the sense that we developed it as a simple proof-of-concept to test our hypothesis: can formal argumentative reasoning enhance LLMs performances? Figure \ref{fig:naive pipeline} depicts the proposed pipeline's high-level operations, which can be delineated as follows:
\begin{enumerate}
    \item\textbf{User prompt.} This component plainly refers to any input provided by the user. The same prompt is fed both to the underlying LLM within the plugin and the model that would provide the final system output.
    \item\textbf{Mistral 7B Instruct.} It indicates the pivotal LLM that drives the argumentation engine and leads to the output reply. In our experiment, we made use of MQInstruct (i.e., the quantizated version of Mistral-7B-Instruct-v0.2).
    \item \textbf{Argument generation.} During this step, the user input is rephrased in order to request that the underlying model produces three short replies to the user prompt and lists three supporting arguments for each generated answer. All the elicited responses will thus compose the set of arguments \emph{AR}.
    \item \textbf{Conflict detection.} At this stage, MQInstruct will be requested to analyse each argument included in AR to record any inconsistency existing among the information they convey. The outcome of this process will comprise the set of attack relations \emph{$\mathcal{C}$}.
    \item \textbf{Argumentation framework.} This element overtly represents the formation of an AF from the previously devised arguments (AR) and their relations ($\mathcal{C}$).
    \item \textbf{ASPARTIX.} So far, the work of the argumentation engine has produced an AF. Now, the ASPARTIX solver has all the required ingredients to compute the \emph{grounded} extension (or \emph{preferred} in case the grounded extension is empty) and its members.
    \item \textbf{Output reply.} The final outcome of the system results from an additional prompt to the LLM with the same initial user input, augmented by the summarized information embedded in the computed acceptable (grounded/preferred) arguments by ASPARTIX. This additional data would guide the model reasoning to achieve a more effective response. 
\end{enumerate}
Being quite minimal, the pipeline (MQArgEng) does not provide any kind of fine-grained parsing of arguments throughout its procedures. This usually entails that `oddly shaped' arguments such as poetical verses, lyrics and similar, mathematical formulae and lines of code, were not properly accounted for. Nonetheless, given the simplicity in recognizing the latter (mostly due to their indentation), we manage to provide heuristics that approximately handle arguments containing lines of code.

To appraise MQArgEng, we conduct an extensive assessment of its and MQInstruct (i.e., the baseline) replies against the MT-Bench. 
\section{Evaluation} 
As is customary for the evaluation of the replies of MT-Bench \cite{zheng2023judging}, we leveraged GPT-4 \cite{openai2023gpt4}\footnote{We made use of the version of GPT-4 available in \url{https://chat.openai.com/} on January 2024.}  as a judge and required it to score the LLMs responses from both MQInstruct and MQArgEng in a range of 1 to 10 according to the \emph{single answer grading} approach. 
When dealing with questions where reference answers were given\footnote{We closely followed the MT-Bench prompts as recorded on  \href{https://huggingface.co/datasets/HuggingFaceH4/mt_bench_prompts}{HuggingFace}. There was only one exception (i.e., question id 103) in the benchmark where we resorted to adding a reference answer since (judge) GPT-4 was failing to recognize that the provided candidates' responses should account for the previous question's logical riddle.} (this mostly involves math, coding and reasoning topics) we opted for the \emph{reference-guided grading}. In particular, we adopted the prompting templates depicted in Figure \ref{fig:exmaples of MT-Bench prompting1} and Figure \ref{fig:exmaples of MT-Bench prompting2}. Notice that we also acted as a human `second marker' to check GPT-4 assessment. That is because it is sometimes the case where the primary judge cannot properly understand the question, erroneously considers wrong right replies (or vice versa), or inconsistently grade them. As such, we required GPT-4 to generate multiple evaluations, and (as a rule of thumb) we decided the official grade was the numerical score that was the mode and appeared at least three times.
This has been also made necessary by the fact that we could not harness the automated MT-Bench evaluation proposed on \href{https://github.com/lm-sys/FastChat/tree/main/fastchat/llm_judge}{FastChat}, given that our pipeline is not an LLM per se. 
\begin{table}
\vspace{0.5cm}
\caption{\footnotesize{Evaluation scores over the MT-Bench.}}
\vspace{0.5cm}
    \centering
    \begin{tabular}{|c|c|c|c|}
    \hline 
    \emph{\textbf{Categories}}  & \textbf{MQInstruct} & \textbf{MQArgEng} & $\mathbf{\Delta}$\\
         \hline
        \emph{Writing} & $8.15$ & $8.05$ & \cellcolor{myred} $\downarrow -0.10$ \\
        \hline
        \emph{Roleplaying} & $6.80$ & $6.75$ & \cellcolor{myred} $\downarrow -0.05$ \\
        \hline
        \emph{Reasoning} & $4.05$& $4.25$& \cellcolor{mygreen}$\uparrow +0.20$\\
        \hline
        \emph{Math} & $2.80$& $2.80$ & $=$ \\
        \hline
        \emph{Coding} & $5.30$& $5.50$ & \cellcolor{mygreen}$\uparrow +0.20$\\
        \hline
        \emph{Extraction} & $5.55$& $5.70$ & \cellcolor{mygreen}$\uparrow +0.15$\\
        \hline
        \emph{STEM} & $7.35$& $7.60$& \cellcolor{mygreen}$\uparrow +0.25$\\
        \hline
        \emph{Humanities} &$7.75$ &$8.10$ & \cellcolor{mygreen}$\uparrow +0.35$\\
        \hline
        \cellcolor{myblue}\emph{\textbf{Average}} & \cellcolor{myblue}$ \mathbf{5.96}$ & \cellcolor{myblue}$\mathbf{6.09}$ & \cellcolor{myblue}$\uparrow \mathbf{+2,18\%}$\\
        \hline
    \end{tabular}
    \label{tab:results}
\end{table}

\begin{figure}[h]
    \centering
\includegraphics[width=0.9\linewidth]{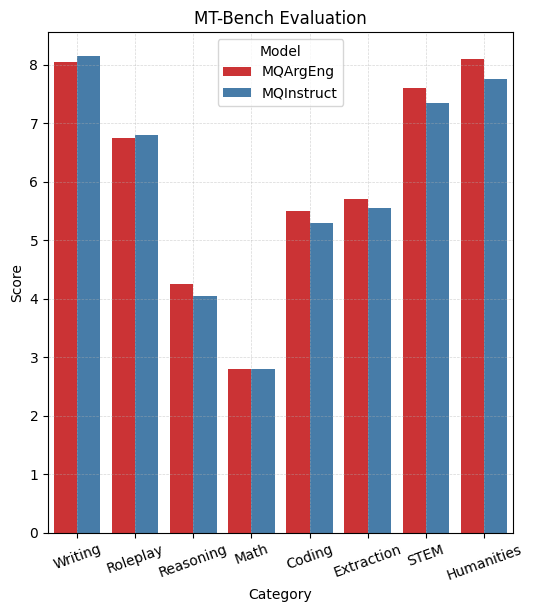}
    \caption{\footnotesize{Graphical rendering of the evaluation scores from Table \ref{tab:results}.}}
    \label{fig:barplot}
    \vspace{1cm}
\end{figure}
\vspace{1cm}

\begin{figure}[h]
    \centering
\includegraphics[width=1\linewidth]{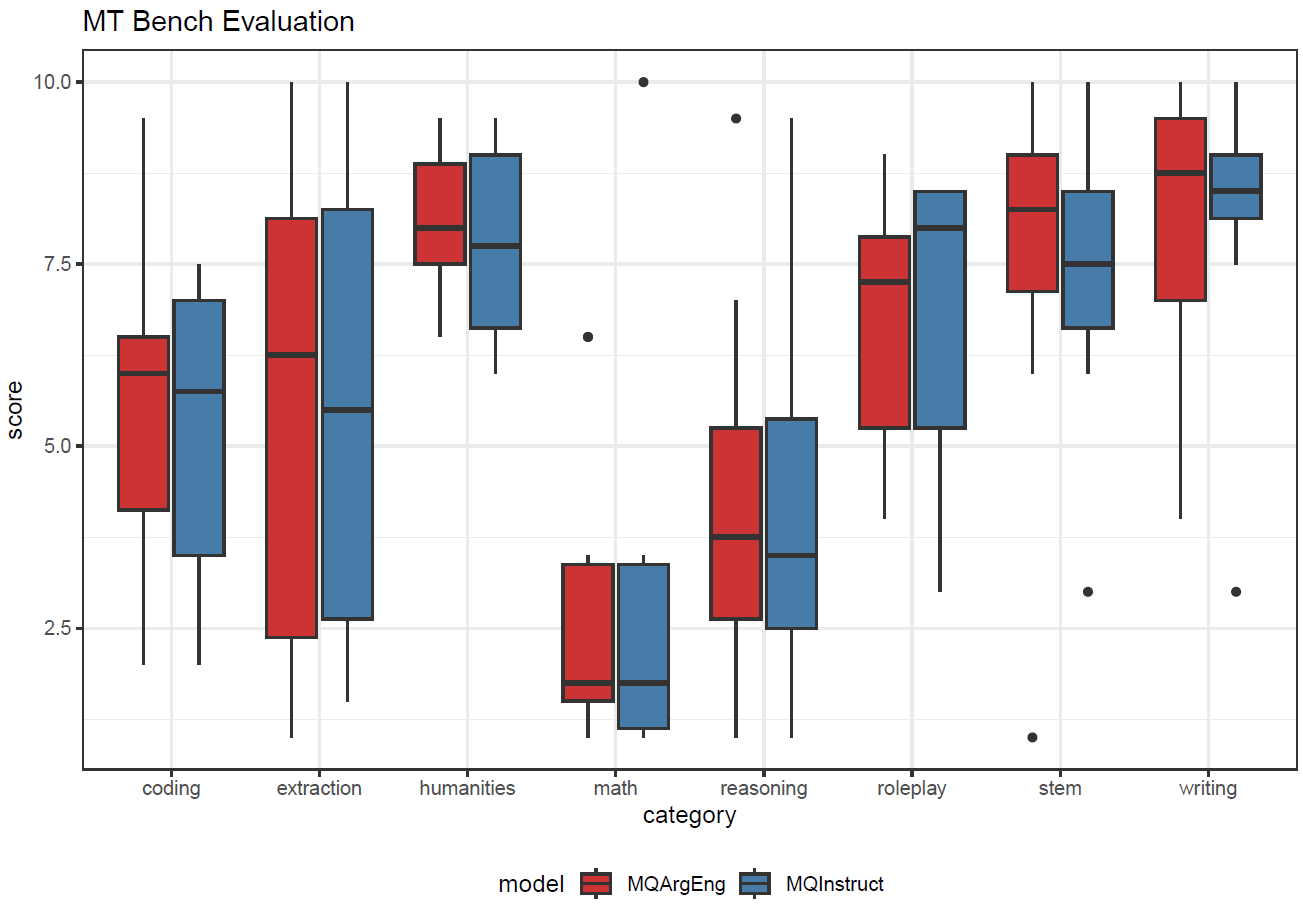}
    \caption{\footnotesize{Data distribution of the evaluation scores from Table \ref{tab:results}.}}
    \label{fig:boxplot}
    \vspace{1cm}
\end{figure}
\section{Discussion}
Before diving into the analysis, we should disclose the curious fact that the total score of MQInstruct does not correspond to the one presented on the current leaderboard\footnote{\url{https://chat.lmsys.org/?leaderboard}.}. This ensues for a series of reasons. For example, the model we leveraged was a quantizated version of Mistral-7b-Instruct v.02, which is supposedly less accurate. Furthermore, we added Zero-shot-CoT to the prompt that, although helpful in the reasoning and maths questions, it is sometimes a disadvantage for writing and roleplaying. Also, consider that our evaluation slightly differed from the one presented in \cite{zheng2023judging} since we introduced a human moderator in the process, and we prompted GPT-4 multiple times for the same question. That being said, our findings are not affected by the difference in score with the leaderboard. We conducted the experiment under the same conditions with both the baseline model (MQInstruct) and the proposed pipeline, and the only metrics we were interested in were the recorded performances and their delta, as shown in Table \ref{tab:results}. Figure \ref{fig:barplot} illustrates how MQArgEng outscored its competitor in 5/8 categories (reasoning, coding, extraction, stem, humanities) whilst tied in the math category of the MT-Bench. These results are further corroborated by Figure \ref{fig:boxplot} where the interquartile range of the data distribution reflects the same improvements of MQArgEng over the baseline, accounting also for the same number of outliers. Overall, given the exploratory stage of the attempted experiment, the scores achieved against the appraised benchmark are quite positive and warrant future research. In particular, the overall $+2,18\%$ is well in scale with the usual improvement granted by LLM research. For instance, the recently released Llama 3 70B-Instruct increased its performance by about $7\%$ compared with Claude 3-Sonnet on the same benchmarks (notice this improvement is the result of a full cycle of pre-training and fine-tuning on over 15T of data, thus definitely more resource intensive than our naive plugin approach \cite{llama3}).  
We can summarize the pros and cons of our findings as follows. 
\subsection{Advantages}
\begin{itemize}
    \item[$\diamond$] As developed herein, an argumentation engine plugin proves to be scalable to any other model (potentially including Small Language Models \cite{slm2024} or even future releases of new LLMs). Granted the generation of multiple different responses and the capability of comparing and detecting inconsistency among the information they embed, there are indeed no restrictions on the underlying model to be leveraged by the pipeline.
    \item[$\diamond$] In addition, there are also no constraints in employing different LLM architectures: designed as an external plugin, an argumentation engine operates regardless of the model structure whether it is a Transformer, the recently introduced Mamba \cite{gu2023mamba} or any other.   
    \item[$\diamond$] Even if the argumentation engine does not outscore the baseline across the board (i.e., writing and roleplaying), it can be extremely useful when deployed for those categories. For example, the plugin could be triggered only when in the presence of such subjects, thus enhancing the LLM output quality, and remain inactive otherwise.
    \item[$\diamond$] The final reply after the injection of the information summarized by the argumentation engine leverages also zero-shot-CoT. It is thus safe to assume that more advanced prompting techniques could be employed, possibly leading to higher performances.
\end{itemize}
\subsection{Limitations} 
\begin{itemize}
    \item[$\diamond$] The parsing of the arguments for further processing along the pipeline was quite straightforward and did not account for nuances in the argument syntax. This mostly affected when arguments were written as lyrics/poems and maths formulae.
    \item[$\diamond$] The overall output heavily relies on the quality of the argument generated by the underlying model, which can slightly vary from one iteration to the other.
    \item[$\diamond$] Similarly, from the underlying model and the available hardware depend the inference speed. For example, making use of a very large LLM (e.g., the open source Grok-1 \cite{grok1}) with inadequate GPU support could suffice for generating output, but the latency between each generation will render the plugin highly ineffective to run.
    \item[$\diamond$] The quality of the replies provided by the argumentation engine within MQArgEng varies between the first and the second sub-questions of the queries of the MT-bench. Indeed, the second sub-questions may sometimes result in lower-score responses. The explanation of this behaviour is related to the initial argument generation from the underlying LLM that may happen to focus on the first half of the prompted queries. This, in turn, drives the engine into reasoning only on one of the sub-questions ending in forcing also the final output of the pipeline to concentrate only on a partial answer.
\end{itemize} 
The poor performance of MQArgEngine on maths-related questions is thus not surprising given the aforementioned limitations (given also that, in general, LLMs have proven to struggle with such topics). The rough parsing of the generated arguments often failed to recognize the math formulae belonging to the unfolding of an equation as a unique element, resulting in arguments consisting of incomplete formulae. Similar consideration holds for the writing and roleplay category, where poetical verses were not properly accounted for. Nonetheless, there were much fewer requests for handling limericks or similar, hence the reason for such a high score in writing subjects compared to maths. In addition, we argue that both writing and roleplay are affected by a somewhat reduced creativity due to the summarized argument injection combined with the requested step-by-step output generation (zero-shot-CoT).   
\section{Related Works} 
As previously anticipated, in an attempt to provide effective solutions to LLMs reasoning shortcomings, several training-free approaches have been proposed in the literature. For example, Self-Consistency Chain, Tree or Graph of Thoughts, respectively, CoT-SC, ToT, and GoT. The limitations of the previously mentioned CoT strategy mostly concern the absence of a procedure to plan or analyse multiple reasoning paths before generating the output, and this is exactly the enhancement yielded by CoT-SC, ToT and GoT. Indeed, Self-Consistency Chain of Thought starts from standard CoT promptings and samples a set of candidate outputs before selecting the answer that is the most consistent among the generated reasoning path \cite{wang2023selfconsistency}.
Tree of Thoughts frames each problem as a search over a tree, where each node is a partial solution \cite{yao2023tree}. Graph of Thoughts, instead, envisages the information generated by an LLM as an arbitrary graph, distilling dependencies between such information units and enhancing reasoning by focusing on the core elements of the network \cite{besta2024graph}. Against these three options, we argue that endowing LLMs' pipelines with a reasoning engine driven by computational argumentation may provide a more intuitive (e.g. grounded on dialectical logic, unlike CoT-SC), cheaper (e.g. less resource-expensive to be implemented, unlike ToT and GoT) and comprehensive alternative (e.g., effective on a variety of topics, unlike the limited use cases showed for ToT and GoT). Argumentative reasoning is particularly suited for models that parse, work and generate natural language. Recall that AFs are graphs whose edges represent paths determining the status of each node. Then, semantically computing an argumentation framework allows planning the most appropriate sequence of `thoughts' (arguments) to achieve the desired result. Such sequences account for divergent information, thus also mimicking and (potentially) outperforming the CCoT (Contrastive Chain of Thought) prompting technique, which generally handles only one contrastive sample at a time \cite{chia2023contrastive}. Unrelated to Chain of Thought, another approach that elicits information from an external engine (i.e., MuJoCo \cite{todorov2012mujoco}) can be found in Google's Mind's Eye \cite{liu2022mind}. Similarly to our pipeline, it adds the output of MuJoCo to the LLM's input and proves how this increases the model reasoning capabilities under the UTOPIA benchmark. Regardless, this procedure requires also the presence of a text-to-code converter to encode data for the engine whose proficiency strictly revolves around physics knowledge. On the other hand, our approach presents a simplified pipeline (harnessing only one LLM), and it is driven by computational argumentation, which ensures augmented capabilities across a broader range of topics (as testified by the MT-Bench evaluation). One last interesting technique to mention is the \emph{step-back prompting} introduced in \cite{zheng2023take}. Indeed, the main idea concerns prompting an LLM to take a step back from the main problem to ask a question about a high-level concept. The answer will then guide the model reasoning about the solution to the original issue. 
Despite the positive outcome, this method requires the generation of step-back questions which are unique for each task in order to retrieve the most relevant facts. On the contrary, our plugin is more flexible and presents a one-size-fits-all design regardless of the circumstances.
\section{Future Directions}
Given the positive outcome of the preliminary study reported herein, we envisage a number of possible research extensions aimed at establishing the plugin's usefulness and increasing its efficiency: 
\begin{itemize}
    \item We believe that employing a stronger underlying LLM (e.g., GPT-4, Claude 3 \cite{claude3}) would improve the overall pipeline output resulting in a higher score on the MT-Bench. That is mostly due to steps a) the generation of arguments and b) the comparisons of arguments to detect conflicting information, both of which solely rely on the capabilities of the leveraged underlying model. Alternatively, another option would be harnessing a separate model specialized to (or fine-tuned for) diversify argument generation and comparison, thus taking charge of steps a) and b). 
    \item Other lines of improvement could originate from the adoption of more advanced prompting techniques to be employed in the final input received by the model or by resorting to better LLMs that would act as judges in the evaluation. This may occur by leveraging the latest cutting-edge model or a fine-tuned version specialized in such an assessment task.   
    \item We argue that better performances could be achieved by also combining together the generated arguments and their supports into single, more complete arguments. Similarly, we posit that a more accurate fine-grained parsing of the `oddly shaped' arguments could improve the overall performance of the pipeline. 
\end{itemize}

\section{Conclusion} 
Is it feasible to integrate computational argumentation within the Large Language Models workflow? Does this yield enhancements in performance?
Motivated by the present shortcomings faced by LLMs with reasoning tasks, in this paper, we have addressed both questions with a positive outcome. In order to do so, we proposed \textit{MQArgEng}, a pipeline that incorporates a computational argumentation engine to guide an LLM output process. We also evaluated it using an experiment to compare its performance to a standard (\textit{MQInstruct}). The results proved how our proposed engine, a simple plugin tool (characterised by having no constraints in terms of the underlying model or their architecture), suffices to increase the MT-Bench scores against the baseline. Such an improvement concerned most of the examined categories, showing particular strength for questions classed as humanities, stem, reasoning and coding whilst slightly failing to achieve equally good scores in questions from categories such as writing and roleplaying. Multiple research directions can lead this study to further boost the plugin's effectiveness by, for example, resorting to state-of-the-art underlying models or employing a more fine-grained parsing of the arguments involved. Overall, we can deem this preliminary experiment as successful and showing promise and warrant further work.





\bibliography{llmarg}

\end{document}